\documentclass[11pt,a4paper]{article}
\usepackage[hyperref]{naaclhlt2019}
\usepackage{times}
\usepackage{latexsym}
\usepackage{amsmath}
\usepackage{pgf}
\usepackage{graphicx}
\usepackage{subcaption}
\usepackage{float}
\usepackage{tikz}
\usepackage{pgfplots}
\usepackage{multirow}
\usepackage{makecell}
\usetikzlibrary{arrows,automata,positioning}
\usepackage{url}
\usepackage{layouts}

\DeclareMathOperator*{\argmin}{arg\,min}
\aclfinalcopy

\title{Context-Aware Cross-Lingual Mapping}
\author{Hanan Aldarmaki$^1$ \and Mona Diab$^{1,2}$\\
    $^1$The George Washington University\\%
    $^2$AWS, Amazon AI\\
    {\tt aldarmaki@gwu.edu,diabmona@amazon.com}
}

\if{false}
\author{Hanan Aldarmaki\\
  The George Washington University  \\
  Washington, DC\\
  {\tt aldarmaki@gwu.edu} \\\And
  Mona Diab\\
  The George Washington University\\
  AWS, Amazon AI\\
  Palo Alto, CA\\
  {\tt  diabmona@amazon.com} \\}

\date{}
\fi

\begin{document}
\maketitle
\begin{abstract}
Cross-lingual word vectors are typically obtained by fitting an orthogonal matrix that maps the entries of a bilingual dictionary from a source to a target vector space. Word vectors, however, are most commonly used for sentence or document-level representations that are calculated as the weighted average of word embeddings. In this paper, we propose an alternative to word-level mapping that better reflects sentence-level cross-lingual similarity. We incorporate context in the transformation matrix by directly mapping the averaged embeddings of aligned sentences in a parallel corpus. We also implement cross-lingual mapping of deep contextualized word embeddings using parallel sentences with word alignments. In our experiments, both approaches resulted in cross-lingual sentence embeddings that outperformed context-independent word mapping in sentence translation retrieval. Furthermore, the sentence-level transformation could be used for word-level mapping without loss in word translation quality.
\end{abstract}

\section{Introduction}
Cross-lingual word vector models aim to embed words from multiple languages into a shared vector space to enable cross-lingual transfer and dictionary expansion \cite{upadhyay2016cross}. One of the most common and effective approaches for obtaining bilingual word embeddings is by fitting a linear transformation matrix on the entries of a bilingual seed dictionary \cite{mikolov2013exploiting}. This approach is versatile and scalable: multilingual embeddings can be obtained by mapping the vector spaces of multiple languages into a shared target language, typically English. In addition, imposing an orthogonality constraint on the mapping ensures that the original pair-wise distances are preserved after the transformation and results in better word translation retrieval \cite{artetxe2016learning,smith2017offline}. 

While word vector spaces tend to be globally consistent across language variations \cite{aldarmaki2018unsupervised}, individual words like homographs with unrelated senses (e.g. `bank', `coast') and phrasal verbs (`stand up', `stand out') are likely to behave less consistently in multilingual vector spaces due to their different usage distributions. Consequently, using such words in the alignment dictionary may result in suboptimal overall mapping. We propose two approaches to counteract this effect by incorporating sentential context in the mapping process without explicit word sense disambiguation or additional linguistic resources. The first approach is based on the recently proposed contextualized embeddings from language models, ELMo \cite{peters2018deep}.  Using a parallel corpus with word-alignments, we extract contextualized embeddings to construct a context-aware dictionary for mapping. The second approach is to learn a transformation between sentence embeddings rather than individual word embeddings. Since these embeddings include context that spans full sentences, we surmise that a mapping learned at this level would be more robust to individual word misalignments. 

We used a constrained set of parallel sentences ranging from one hundred to a million sentences for alignment. We then evaluated the resultant mappings on sentence translation retrieval among English, Spanish, and German as test languages. Our results show that context-aware mappings significantly outperform context-independent cross-lingual word mappings using reasonably-sized parallel corpora, particularly when using contextualized word embeddings. In addition, when averaging static word embeddings, the sentence-level mapping can still be used for word-level mapping without loss in word translation quality. 

\section{Related Work}
For cross-lingual alignment, we follow the popular approach of fitting a linear transformation matrix between word vector spaces that are independently trained for each language. Aligning mono-lingual word vector spaces using a seed dictionary was originally proposed in \newcite{mikolov2013exploiting}.  In \newcite{artetxe2016learning} and \newcite{smith2017offline}, it was shown that imposing an orthogonality constraint on the transformation leads to better word translation quality. Recently, contextualized word embeddings were proposed, where a sequential neural network is trained as a language model and then used to extract context-sensitive word representations from the hidden states \cite{peters2018deep}. We use parallel text in order to align independently-trained contextualized embeddings across languages. \newcite{schuster2019cross} independently proposed a cross-lingual alignment approach for contextualized embeddings without the use of parallel text. 

\section{Approach}
\subsection{Orthogonal Bilingual Mapping} \label{orth}
Given a dictionary of source to target pairs $\langle x, y \rangle$ and matrix representations $X$ and $Y$ whose columns are vector representations of the corresponding dictionary items, we seek to find an orthogonal transformation matrix $R$ that minimizes the distances between the transformed vectors in $RX$ and $Y$. Formally,

\begin{equation}
R=\argmin_{\hat  R} \lVert \hat RX-Y \rVert  \;\;\;  \text{s. t.} \;\; \hat R^T \hat R = I
\end{equation}

\noindent where  $\lVert . \rVert$ denotes the Frobenius norm.  The orthogonality constraint ensures that pair-wise distances in the original source vector space are preserved after the transformation. As shown in \cite{schonemann1966generalized}, the solution can be found by singular value decomposition of $YX^T$

\begin{equation*}
YX^T=U\Sigma V^T
\end{equation*}
\noindent Then,
\begin{equation}
R=UV^T
\end{equation}

The resultant transformation, $R$, can then be used to transform additional vectors in the source vector space. The quality of the transformation depends on the size and accuracy of the initial dictionary, and it is typically evaluated on word translation precision using nearest neighbor search \cite{smith2017offline}.

\subsection{Mapping of Contextualized Embeddings}
Word embeddings in a given language tend to have similar structures as their translations in a target language \cite{aldarmaki2018unsupervised}, which enables orthogonal mappings of word vector spaces to generalize well across various languages. However, items in bilingual dictionaries typically refer to specific word senses. In a given dictionary pair, the source word may have multiple senses that are not consistent with its aligned target translation (and vise versa), which could result in suboptimal global mappings. Intuitively, better mappings could be obtained using sense-disambiguated word embeddings, which could be approximated from context. ELMo (Embeddings from Language Models) is a recently-proposed deep model for obtaining contextualized word embeddings, which are calculated as the hidden states of a bi-LSTM network trained as a language model \cite{peters2018deep}. The network can be used in lieu of static word embeddings within other models, which yields better performance in a range of tasks, including word sense disambiguation. Sentence embeddings can be obtained from ELMo by averaging the contextualized word embeddings \cite{perone2018evaluation}. 

Since ELMo generates dynamic, context-dependent vectors, we cannot use a simple word-level dictionary to map the model across languages. Instead, we use a parallel corpus with word alignments, i.e using an IBM Model \cite{brown1993mathematics}, to extract a dynamic dictionary of aligned contextualized word embeddings. Depending on the size of the parallel corpus, a large dictionary can be extracted to learn an orthogonal mapping as described in Section \ref{orth}, which is then applied post-hoc on newly generated contextualized embeddings. 

 \subsection{Sentence-Level Mapping} 
An alternative general approach for obtaining a context-aware mapping is to learn sentence-level transformations. Intuitively, a sentence is less ambiguous than stand-alone words since the words are interpreted within a specific context, so a mapping learned at the sentence-level is likely to be less sensitive to individual word inconsistencies. Therefore, we learn the mapping as described in Section \ref{orth} using a dictionary of aligned sentence embeddings.  Over a large parallel corpus, the aggregate mapping can yield a more optimal global solution compared to word-level mapping. This approach can be applied using any model capable of generating monolingual sentence embeddings. In this work, we use the average of word vectors in each sentence, where the word vectors are either static or contextualized. For inference, monolingual sentence embeddings are generated first, then mapped to the target space using the sentence-level transformation matrix.\footnote{Since we use vector averaging, it doesn't matter whether we apply the learned transformation to the word embeddings before averaging, or to the sentence embeddings after averaging. } 

\section{Experiments}

We used skip-gram with subword information, i.e FastText \cite{bojanowski2017enriching}, for the static word embeddings, and ELMo for contextualized word embeddings. Sentence embeddings were calculated from ELMo as the arithmetic average of the contextualized embeddings \footnote{We found that using the arithmetic average for ELMo yields better results than weighted averaging.}. For FastText, we applied weighted averaging using smooth inverse frequency \cite{arora2017asimple}, which works better for sentence similarity compared to other averaging schemes \cite{aldarmaki2018evaluation}. 

\subsection{Data and Processing}

We trained and aligned all models using the same monolingual and parallel datasets. For monolingual training, we used the 1 Billion Word benchmark \cite{chelba2014one} for English, and equivalent subsets of $\sim$400 million tokens from WMT'13 \cite{bojar2013findings} news crawl data. We trained monolingual ELMo and FastText with default parameters. We used the WMT'13 common-crawl data for cross-lingual mapping, and the WMT'13 test sets for evaluating sentence translation retrieval. For all datasets, the only preprocessing we performed was tokenization.

\subsection{Evaluation Framework}

We evaluated the cross-lingual mapping approaches on sentence translation retrieval, where we calculate the accuracy of retrieving the correct translation from the target side of a test parallel corpus using nearest neighbor search with cosine similarity.  To assess the minimum bilingual data requirements of each approach and measure how the various models respond to additional data, we split the training parallel corpus into smaller subsets of increasing sizes, starting from 100 to a million sentences (we double the size at each step). Data splits and evaluation scripts are available at https://github.com/h-aldarmaki/sent\_translation\_retrieval.

\subsection{Alignment Schemes}

For ELMo, word embeddings need to be calculated from context, so we extracted a dictionary of contextualized words from the parallel corpora by first applying word-level alignments using Fast\_Align \cite{dyer2013simple}. We then calculated the contextualized embeddings for source and target sentences, and extracted a dictionary from the aligned words that have a one-to-one alignment (i.e. we excluded phrasal alignments). Since this can result in a very large dictionary, we capped the number of dictionary words at 1M for efficiency. 
For a fair comparison with FastText word-level mapping, we extracted a dictionary from word alignment probabilities using the same parallel sets. For each word in the source language, we extracted its translation as the word with the maximum alignment probability if the maximum was unique\footnote{Using other dictionary pairs generally resulted in lower performance.}. As a baseline, we used static dictionaries from \cite{conneau2017word} to obtain word-level mappings \texttt{(dict)}. All alignments were performed from the source languages to English.

\subsubsection{Results}

\begin{figure} [H]
\hspace{-0.2cm}
\begin{subfigure}{0.238\textwidth}
\centering
\small
 \begin{tikzpicture}[font=\sffamily, scale=0.80, transform shape]
\begin{semilogyaxis}[title=es $\rightarrow$ en ,
                ymode=normal,
                ymin=0,
                ymax=100,
                xmin=0,
                ytick={20, 50, 80, 100},
                yticklabels={.2, .5, .8, 1},
                xtick={1, 10,100, 1000},
                xticklabels={1K,10K,100K, 1M}, %
                xmode=log, 
                log basis x={2},
                axis x line=bottom,
                axis y line=left,
                axis line style=-,
                x=0.3cm, y=0.03cm,
                ylabel = NN accuracy,
                compat=newest,
                y label style={at={(-0.1,0.5)}},
                ]
\addplot[blue, dotted, mark=*,mark options={solid, scale=0.7}] table [x=size, y=elmo_avg, col sep=space] {es_en.nn};
\addplot[brown,mark=o,mark options={solid, scale=0.7}] table [x=size, y=elmo_sent, col sep=space] {es_en.nn};
\addplot[red, dotted, mark=x,mark options={solid, scale=1, fill=black}] table [x=size, y=siskip_sif, col sep=space] {es_en.nn};
\addplot[green,mark=none,mark options={solid, scale=1}] table [x=size, y=siskip_sent, col sep=space] {es_en.nn};
\addplot [dashed] table [x=size, y=muse_sif, col sep=space] {es_en.nn};
\end{semilogyaxis}
\end{tikzpicture}
\end{subfigure}%
\hspace{0.1cm}
\begin{subfigure}{0.238\textwidth}
\centering
\small

 \begin{tikzpicture}[font=\sffamily, scale=0.80, transform shape]
\begin{semilogyaxis}[title=de $\rightarrow$ en ,
                ymode=normal,
                ymin=0,
                ymax=100,
                xmin=0,
                 ytick={20, 50, 80, 100},
                yticklabels={.2, .5, .8, 1},
                xtick={1, 10,100, 1000},
                xticklabels={1K,10K,100K, 1M}, %
                xmode=log, 
                log basis x={2},
                axis x line=bottom,
                axis y line=left,
                axis line style=-,
                x=0.3cm, y=0.03cm,
                ]
\addplot[blue, dotted, mark=*,mark options={solid, scale=0.7}] table [x=size, y=elmo_avg, col sep=space] {de_en.nn};
\addplot[brown,mark=o,mark options={solid, scale=0.7}] table [x=size, y=elmo_sent, col sep=space] {de_en.nn};
\addplot[red, dotted, mark=x,mark options={solid, scale=1, fill=black}] table [x=size, y=siskip_sif, col sep=space] {de_en.nn};
\addplot[green,mark=none,mark options={solid, scale=1}] table [x=size, y=siskip_sent, col sep=space] {de_en.nn};
\addplot [dashed] table [x=size, y=muse_sif, col sep=space] {de_en.nn};
\end{semilogyaxis}
\end{tikzpicture}
\end{subfigure}\\%

\vspace{0.5cm}
\hspace{-0.2cm}
\begin{subfigure}{0.238\textwidth}
\centering
\small
 \begin{tikzpicture}[font=\sffamily, scale=0.80, transform shape]
\begin{semilogyaxis}[title=en $\rightarrow$ es ,
                 title style={at={(0.5,1.0)},anchor=north,yshift=-0.1},
                ymode=normal,
                ymin=0,
                ymax=100,
                xmin=0,
                 ytick={20, 50, 80, 100},
                yticklabels={.2, .5, .8, 1},
                xtick={1, 10,100, 1000},
                xticklabels={1K,10K,100K, 1M}, %
                xmode=log, 
                log basis x={2},
                axis x line=bottom,
                axis y line=left,
                axis line style=-,
                x=0.3cm, y=0.03cm,
                ylabel = NN accuracy,
                compat=newest,
                y label style={at={(-0.1,0.5)}},
                ]
\addplot[blue, dotted, mark=*,mark options={solid, scale=0.7}] table [x=size, y=elmo_avg, col sep=space] {en_es.nn};
\addplot[brown,mark=o,mark options={solid, scale=0.7}] table [x=size, y=elmo_sent, col sep=space] {en_es.nn};
\addplot[red, dotted, mark=x,mark options={solid, scale=1, fill=black}] table [x=size, y=siskip_sif, col sep=space] {en_es.nn};
\addplot[green,mark=none,mark options={solid, scale=1}] table [x=size, y=siskip_sent, col sep=space] {en_es.nn};
\addplot [dashed] table [x=size, y=muse_sif, col sep=space] {en_es.nn};
\end{semilogyaxis}
\end{tikzpicture}
\end{subfigure}%
\hspace{0.1cm}
\begin{subfigure}{0.238\textwidth}
\centering
\small

 \begin{tikzpicture}[font=\sffamily, scale=0.80, transform shape]
\begin{semilogyaxis}[title=en $\rightarrow$ de ,
                ymode=normal,
                 title style={at={(0.5,1.0)},anchor=north,yshift=-0.1},
                ymin=0,
                ymax=100,
                xmin=0,
                 ytick={20, 50, 80, 100},
                yticklabels={.2, .5, .8, 1},
                xtick={1, 10,100, 1000},
                xticklabels={1K,10K,100K, 1M}, %
                xmode=log, 
                log basis x={2},
                axis x line=bottom,
                axis y line=left,
                axis line style=-,
                x=0.3cm, y=0.03cm,
                ]
\addplot[blue, dotted, mark=*,mark options={solid, scale=0.7}] table [x=size, y=elmo_avg, col sep=space] {en_de.nn};
\addplot[brown,mark=o,mark options={solid, scale=0.7}] table [x=size, y=elmo_sent, col sep=space] {en_de.nn};
\addplot[red, dotted, mark=x,mark options={solid, scale=1, fill=black}] table [x=size, y=siskip_sif, col sep=space] {en_de.nn};
\addplot[green,mark=none,mark options={solid, scale=1}] table [x=size, y=siskip_sent, col sep=space] {en_de.nn};
\addplot [dashed] table [x=size, y=muse_sif, col sep=space] {en_de.nn};
\end{semilogyaxis}
\end{tikzpicture}
\end{subfigure}%

\vspace{0.5cm}
\hspace{-0.2cm}
\begin{subfigure}{0.238\textwidth}
\centering
\small
 \begin{tikzpicture}[font=\sffamily, scale=0.80, transform shape]
\begin{semilogyaxis}[title=es $\rightarrow$ de ,
                 title style={at={(0.5,1.0)},anchor=north,yshift=-0.1},
                ymode=normal,
                ymin=0,
                ymax=100,
                xmin=0,
                 ytick={20, 50, 80, 100},
                yticklabels={.2, .5, .8, 1},
                xtick={1, 10,100, 1000},
                xticklabels={1K,10K,100K, 1M}, %
                xmode=log, 
                log basis x={2},
                axis x line=bottom,
                axis y line=left,
                axis line style=-,
                x=0.3cm, y=0.03cm,
                ylabel = NN accuracy,
                compat=newest,
                y label style={at={(-0.1,0.5)}},
                xlabel = \# parallel sentences
                ]

\addplot[blue, dotted, mark=*,mark options={solid, scale=0.7}] table [x=size, y=elmo_avg, col sep=space] {es_de.nn};
\addplot[brown,mark=o,mark options={solid, scale=0.7}] table [x=size, y=elmo_sent, col sep=space] {es_de.nn};
\addplot[red, dotted, mark=x,mark options={solid, scale=1, fill=black}] table [x=size, y=siskip_sif, col sep=space] {es_de.nn};
\addplot[green,mark=none,mark options={solid, scale=1}] table [x=size, y=siskip_sent, col sep=space] {es_de.nn};
\addplot [dashed] table [x=size, y=muse_sif, col sep=space] {es_de.nn};
\end{semilogyaxis}
\end{tikzpicture}
\end{subfigure}%
\hspace{0.1cm}
\begin{subfigure}{0.238\textwidth}
\centering
\small

 \begin{tikzpicture}[font=\sffamily, scale=0.80, transform shape]
\begin{semilogyaxis}[title=de $\rightarrow$ es ,
                ymode=normal,
                 title style={at={(0.5,1.0)},anchor=north,yshift=-0.1},
                ymin=0,
                ymax=100,
                xmin=0,
                 ytick={20, 50, 80, 100},
                yticklabels={.2, .5, .8, 1},
                xtick={1, 10,100, 1000},
                xticklabels={1K,10K,100K, 1M}, %
                xmode=log, 
                log basis x={2},
                axis x line=bottom,
                axis y line=left,
                axis line style=-,
                x=0.3cm, y=0.03cm,
                xlabel = \# parallel sentences
                ]
\addplot[blue, dotted, mark=*,mark options={solid, scale=0.7}] table [x=size, y=elmo_avg, col sep=space] {de_es.nn};
\addplot[brown,mark=o,mark options={solid, scale=0.7}] table [x=size, y=elmo_sent, col sep=space] {de_es.nn};
\addplot[red, dotted, mark=x,mark options={solid, scale=1, fill=black}] table [x=size, y=siskip_sif, col sep=space] {de_es.nn};
\addplot[green,mark=none,mark options={solid, scale=1}] table [x=size, y=siskip_sent, col sep=space] {de_es.nn};
\addplot [dashed] table [x=size, y=muse_sif, col sep=space] {de_es.nn};
\end{semilogyaxis}
\end{tikzpicture}
\end{subfigure}\\ 
\vspace{1.5pt}
\begin{subfigure}{0.238\textwidth}
\centering
\small
 \begin{tikzpicture}[font=\sffamily, scale=0.8, transform shape]
\begin{semilogyaxis}[title=Average ,
                ymode=normal,
                 title style={at={(0.5,1.0)},anchor=north,yshift=-0.1},
                ymin=0,
                ymax=100,
                xmin=0,
                 ytick={20, 50, 80, 100},
                yticklabels={.2, .5, .8, 1},
                xtick={1, 10,100, 1000},
                xticklabels={1K,10K,100K, 1M}, %
                xmode=log, 
                log basis x={2},
                axis x line=bottom,
                axis y line=left,
                axis line style=-,
                x=0.3cm, y=0.03cm,
                xlabel = \# parallel sentences
                ]
\addplot[blue, dotted, mark=*,mark options={solid, scale=0.7}] table [x=size, y=elmo_avg, col sep=space] {avg.nn};
\addplot[brown,mark=o,mark options={solid, scale=0.7}] table [x=size, y=elmo_sent, col sep=space] {avg.nn};
\addplot[red, dotted, mark=x,mark options={solid, scale=1, fill=black}] table [x=size, y=siskip_sif, col sep=space] {avg.nn};
\addplot[green,mark=none,mark options={solid, scale=1}] table [x=size, y=siskip_sent, col sep=space] {avg.nn};
\addplot [dashed] table [x=size, y=muse_sif, col sep=space] {avg.nn};
\end{semilogyaxis}
\end{tikzpicture}
\end{subfigure}
\begin{subfigure}{0.21\textwidth}
\small
 \begin{tikzpicture}[font=\sffamily, scale=0.750, transform shape]
\begin{axis}[
    hide axis,
    xmin=0,
    xmax=0.5,
    ymin=0,
    ymax=0.4,
    legend style={at={(0.3,0.8)},anchor=north east,draw=white!15!black,legend cell align=left},
    ]
\addlegendimage{blue, dotted, mark=*,mark options={solid, scale=0.7}}
\addlegendentry{ELMo \hspace{9pt}\texttt{(word)} 82.23\%};
\addlegendimage{brown,mark=o,mark options={solid, scale=0.7}}
\addlegendentry{ELMo \hspace{9pt}\texttt{(sent)} 84.03\%};
\addlegendimage{red, dotted, mark=x,mark options={solid, scale=1, fill=black}} 
\addlegendentry{FastText \texttt{(word)} 74.00\%};
\addlegendimage{green,mark=none,mark options={solid, scale=1}}
\addlegendentry{FastText \texttt{(sent)} 76.92\%};
\addlegendimage {dashed}
\addlegendentry{FastText \texttt{(dict)} 69.04\%};
\end{axis}
\end{tikzpicture}
\end{subfigure}
\begin{minipage}{0.48\textwidth}
\centering
\caption{Nearest neighbor sentence translation accuracy as a function of (log) parallel corpus size. \texttt{(word)} refers  to word-level mapping, \texttt{(sent)} to sentence-level mapping, and \texttt{(dict)} refers to the baseline (using a static dictionary for mapping). The legend shows the average accuracies of each model using 1M parallel sentences.} \label{fig:nn_acc}
\end{minipage}\hfill
\end{figure}
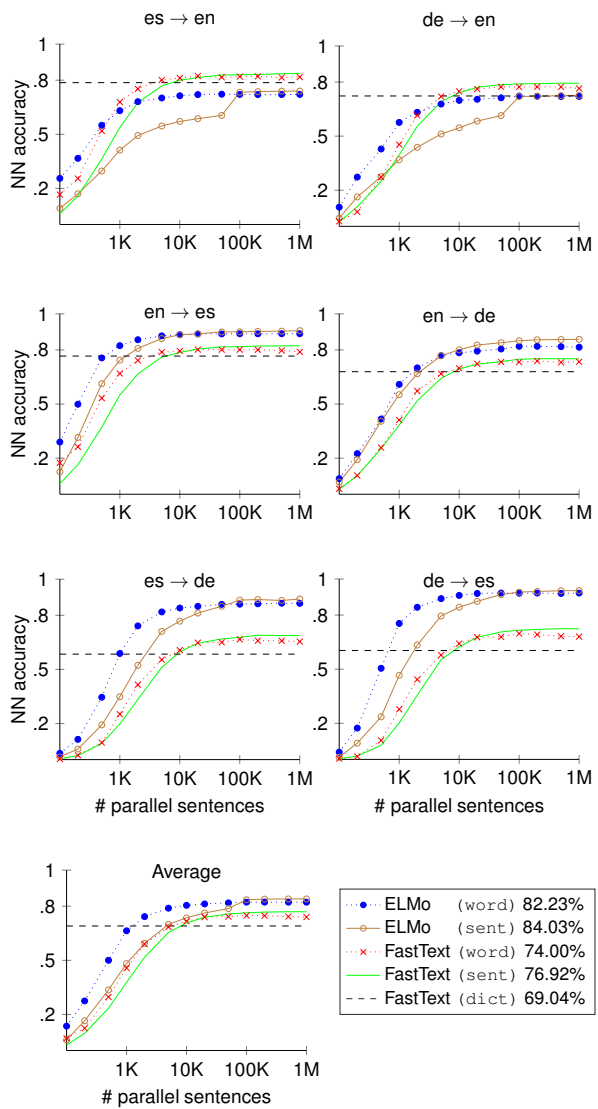

Sentence translation retrieval results in all language directions are shown in Figure \ref{fig:nn_acc}  (note the x-axis denotes the size of the alignment corpus in log scale). The arrows indicate the translation direction from source to target, with $en$ for English, $es$ for Spanish, and $de$ for German. For clarity, the legend shows the average accuracies in the final step (1M). 

Overall, ELMo word alignment resulted in the highest sentence translation retrieval accuracies, even with small amounts of training data; it exceeded the static dictionary baseline at around 2K parallel sentences. Sentence-level mapping outperformed word-level mapping only when additional parallel data were used (over 50K sentences). 

With 1M sentences, sentence-level mapping of FastText yielded an increase of $\sim$3\% in all directions. Sentence-level ELMo underperformed in the $\rightarrow en$  directions until we used 100K sentences, where we observed a sharp increase in accuracy compared to the previous step of 50K sentences.
For ELMo, we note particular improvements in zero-shot translation retrieval between the source languages: $es$ and $de$, where ELMo-based models performed much higher than FastText. The opposite is true for the $\rightarrow en$ directions, although the difference is not as notable.  This is an interesting observation and may indicate that contextualized dictionaries result in a more balanced mapping, while context-independent embeddings overfit the mapping to the specific direction used for alignment.

\subsubsection{Word-level Evaluation}

Cross-lingual word embeddings are typically evaluated in word-translation retrieval: the precision of correctly retrieving a translation from the vocabulary of another language. Since this is a context-free task, we evaluated the performance of static word embeddings, FastText, using word vs. sentence mapping (with 1M parallel sentences). The transformation matrix learned at the sentence level is used to transform the word embeddings. 

\begin{table} [t]
\centering
\scalebox{0.8}{
\begin{tabular}{|p{2cm}|p{1cm}|c|c|}
\hline
\multicolumn{2}{|c}{\multirow{2}{*}{ Language pair} } &  \multicolumn{2}{|c|}{Mapping level}  \\
\cline{3-4}
\multicolumn{2}{|c|}{} & word & sentence \\
 \hline
  \multicolumn{4}{|l|}{\textbf{From source language to en:}} \\
  \hline
 \multirow{2}{*}{ es-en} & k=1 & \textbf{56.46}	 &54.43\\
\cline{2-4}
    & k=5    & \textbf{70.93} &	68.97\\
    \hline
 \multirow{2}{*}{ de-en} & k=1 & \textbf{50.00}]	& 47.85\\
\cline{2-4}
    & k=5    & \textbf{63.45}	& 62.69\\
\Xhline{6\arrayrulewidth}
  \multicolumn{4}{|l|}{\textbf{From en to source language:}} \\
  \hline
 \multirow{2}{*}{en-es} & k=1 &56.98 &	\textbf{57.52}\\
\cline{2-4}
    & k=5    & \textbf{72.68} &	72.15\\
      \hline
 \multirow{2}{*}{en-de} & k=1 &  42.32 &	\textbf{43.27}\\
\cline{2-4}
    & k=5    & \textbf{63.99} &	62.84\\
\Xhline{6\arrayrulewidth}
  \multicolumn{4}{|l|}{\textbf{Translation between source languages:}} \\
  \hline
 \multirow{2}{*}{de-es} & k=1 & 36.14	 & \textbf{37.07}\\
\cline{2-4}
    & k=5    & 53.72	& \textbf{54.85}\\
      \hline
 \multirow{2}{*}{es-de} & k=1 & 31.55	 & \textbf{34.22}\\
\cline{2-4}
    & k=5    & 51.37	& \textbf{52.07}\\
\Xhline{6\arrayrulewidth}
 \multirow{2}{*}{Average} & k=1 & 45.58	& \textbf{45.73}\\
\cline{2-4}
    & k=5    & \textbf{62.69}	& 62.26\\
      \hline
\end{tabular}
}
\caption{Word translation precision at $k$ (\%) using $k$ nearest neighbor search, with $k \in \{1,5\}$. } \label{tab:1}
\end{table}
\begin{table} [t]
\centering
\scalebox{0.8}{
\begin{tabular}{|p{2.5cm}|c|c|}
\hline
\multirow{2}{*}{ Language pair} &  \multicolumn{2}{|c|}{Mapping level}  \\
\cline{2-3}
 & word & sentence \\
 \hline
en-es  & 0.6280	& \textbf{0.6362}\\
\hline
en-de    & \textbf{0.6480} &	0.6476\\
    \hline
es-de  & 0.6349	& \textbf{0.6383}\\
\hline
Average    & 0.6370 & \textbf{0.6407}\\
      \hline
\end{tabular}
}
\caption{The harmonic mean of Pearson and Spearman correlations with human judgment on the SemEval'17 cross-lingual word similarity task. } \label{tab:2}
\end{table}

We used the dictionaries from \cite{conneau2017word}. We also evaluated on the SemEval'17 cross-lingual word similarity task \cite{camacho2017semeval}, which is measured using the average of Pearson and Spearman correlation coefficients against human judgements. As shown in Tables \ref{tab:1} and \ref{tab:2}, the mapping learned at the sentence-level yields equivalent performance to word-level mapping. While word-level mapping was slightly better in translating from source languages (German and Spanish) to English, the sentence-level mapping was better when translating between the source languages. In the word similarity task, sentence-level mappings performed slightly better in two out of the three cases. Overall, the performance of both models are comparable, which indicates that a single transformation matrix learned at the sentence-level can be used for both word and sentence-level tasks. 

\section{Discussion and Conclusions}

We introduced alternatives to the popular word mapping approach that incorporate context in the mapping process. Given parallel corpora, context-aware mappings were learned by mapping aligned contextualized word embeddings or directly mapping the parallel sentence embeddings. Experimental results showed significant gains in sentence translation retrieval using contextualized mappings compared to context-independent word mapping. While word-level mappings worked better with smaller parallel corpora, the performance of sentence-level mapping continued to increase with additional data until it outperformed word-level mapping. In future work, we will explore the viability of the sentence mapping approach on other sentence embedding models.

\bibliography{refs}
\bibliographystyle{acl_natbib}

\
\end{document}